\documentclass[runningheads]{llncs}
\usepackage{comment}
\usepackage{graphicx}
\usepackage{amsmath}
\usepackage{cite}

\usepackage{xcolor}
\usepackage{hyperref}
\hypersetup{
    colorlinks=true,
    linkcolor=blue,
    urlcolor=blue,
}
\newcommand{\head}[1]{\textbf{\underline{#1}} \\}

\begin{document}
\title{HOW MUCH CAN ChatGPT REALLY HELP COMPUTATIONAL BIOLOGISTS IN PROGRAMMING?}
\titlerunning{ChatGPT in Bioinformatics Coding}
% If the paper title is too long for the running head, you can set
% an abbreviated paper title here
%
\author{Chowdhury Rafeed Rahman \\
Limsoon Wong
}
\authorrunning{Rafeed}
% First names are abbreviated in the running head.
% If there are more than two authors, 'et al.' is used.
%
\institute{National University of Singapore \\
\email{e0823054@u.nus.edu}
}
\maketitle              % typeset the header of the contribution
\begin{abstract}
ChatGPT, a recently developed product by openAI, is successfully leaving its mark as a multi-purpose natural language based chatbot. In this paper, we are more interested in analyzing its potential in the field of computational biology. A major share of work done by computational biologists these days involve coding up bioinformatics algorithms, analyzing data, creating pipelining scripts and even machine learning modeling and feature extraction. This paper focuses on the potential influence (both positive and negative) of ChatGPT in the mentioned aspects with illustrative examples from different perspectives. Compared to other fields of computer science, computational biology has - (1) less coding resources, (2) more sensitivity and bias issues (deals with medical data) and (3) more necessity of coding assistance (people from diverse background come to this field). Keeping such issues in mind, we cover use cases such as code writing, reviewing, debugging, converting, refactoring and pipelining using ChatGPT from the perspective of computational biologists in this paper. 
\end{abstract}
\section{Introduction}
ChatGPT is a state-of-the-art chatbot capable of conversation in natural language. It was first officially released on November 2022 by openAI trained on terabytes of diverse text data from up until September 2021. Since then, ChatGPT has been influencing important domains such as education, internet searching, coding, job efficiency, research, social interaction and many others. The number of active users of ChatGPT surpassed 100 million in the first two months of its release \footnote{https://dailyinvestor.com/world/8520/chatgptbreaks-record-with-100-million-users-and-investors-come-flocking/}. Some of the other popular openAI tools are DALL-E (text to image generation) \cite{ramesh2021zero} and Codex (assisting in coding tasks) \cite{chen2021evaluating} which are also being heavily used by people for their respective use cases. Recently, Meta and Google have also released their own chatbot named LLaMa \cite{touvron2023llama} and Bard \cite{manyika2023overview}, respectively.  

The basis of ChatGPT is the Transformer encoder-decoder architecture \cite{vaswani2017attention} built based on the attention mechanism. The concept of language modeling based auto-regressive unsupervised pretraining made it possible to use large amount of unlabeled data to enrich such models \cite{radford2018improving}. Although during 2018-2020, masked pretraining based Transformers started becoming popular because of the training efficiency \cite{devlin2018bert}, it was gradually proven that generational pretraining enables the models perform better in zero shot (no task-specific labeled data required) and few shot (very few labeled data required) learning tasks \cite{radford2019language,brown2020language}. ChatGPT is in essence the GPT-3.5 model \cite{ye2023comprehensive} which evolved from GPT-1, GPT-2 and GPT-3 \cite{brown2020language}. The evolution of GPT-1 to GPT-3 is mainly attributed to larger models and larger training corpus. GPT-3.5 is the result of further supervised fine-tuning of GPT-3 model along with the application of reinforcement learning with human feedback \cite{ouyang2022training}.  

\begin{figure}[!htb]
    \centering
    \includegraphics[width=1.0\textwidth]{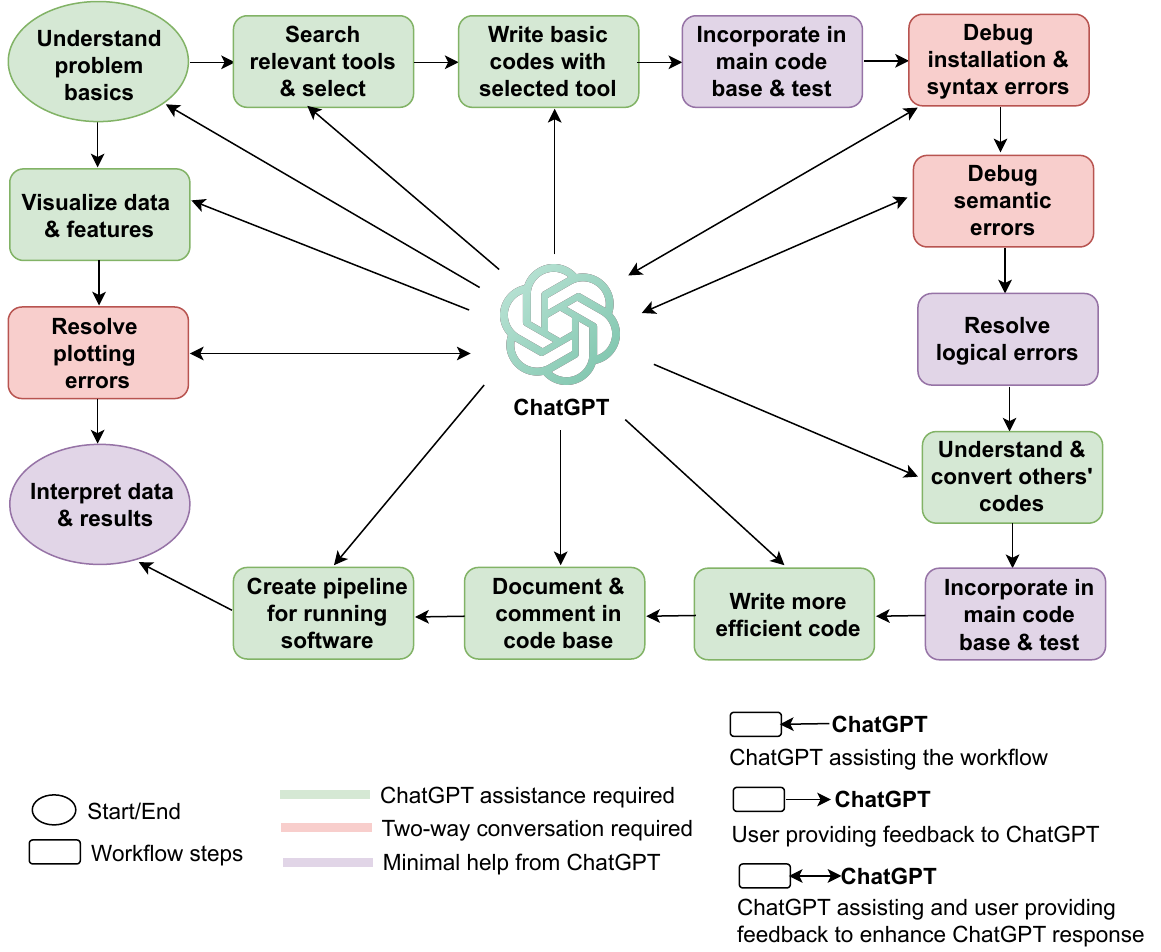}
    \caption{Overview of ChatGPT contributing in computational biology programming workflow}
    \label{fig:overview}
\end{figure}

With the recent diverse usage of ChatGPT, researchers have started investigating this tool from different perspectives. Zhang et al. performed a detail survey on ChatGPT in terms of its potential in artificial general intelligence \cite{zhang2023one}. Sobania et al. on the other hand analyzed its bug fixing capability \cite{sobania2023analysis}. Borji et al. focused on general failure scenarios of ChatGPT such as factual, logical and reasoning error \cite{borji2023categorical}. Another recent paper only focused on the limited mathematical capability of ChatGPT \cite{frieder2023mathematical}. A recent research \cite{lubiana2023ten} has focused on some important tips regarding how ChatGPT can be utilized by computational biologists overall. Another recent research \cite{piccolo2023many} has focused on ChatGPT's performance on bioinformatics coding assignment solving. In this paper, we elaborate the potential use cases and limitations of ChatGPT from computational biology coding perspective. We discuss different aspects such as code generation, reviewing, documentation, debugging, explanation, conversion, concept understanding and data visualization. An overview of how ChatGPT can be used in a computational biology project coding workflow is provided in Figure \ref{fig:overview}. We hope that the real life examples provided in this paper will help users be aware of the potential issues of ChatGPT while using this influential tool. The records of our interaction with ChatGPT mentioned in different parts of the paper have been provided in the last section.

\section{Typical Knowledge Based Question Answering}
Working with a new tool/file format/library/algorithm for the first time can be challenging. We start by asking ChatGPT a couple of questions about Samtools \cite{li2009sequence} and Pysam \footnote{https://github.com/pysam-developers/pysam} (used for manipulating raw BAM and SAM files which are obtained after aligning raw reads to a reference genome). We asked about their use cases and the circumstances when one is preferred over the other. ChatGPT gave us accurate and detailed information (Record S1). We then asked it about the SWISS-PROT database  \cite{bairoch2000swiss} which is used for protein sequence extraction according to their different properties, and got a nice introduction and diverse use cases from ChatGPT (Record S2). While asking such questions, one needs to remember that ChatGPT is a model which was last trained on data of September 2021. So, it cannot provide you with appropriate information if you ask it about recent knowledge based affairs. When asked about the recent cost of 0.1X whole genome sequencing (WGS), it will tell you to go through relevant websites for the most updated information. Also we asked it about the latest version of PSI BLAST tool  \cite{altschul1997gapped} (used for protein sequence similarity searching). It provided us information of an older version (Record S3). ChatGPT can hallucinate false details when answering relatively uncommon descriptive questions. These details are hard to catch unless one is an expert in that relevant field. We asked it about the contribution of clonal hematopoiesis (CH) \cite{chan2020clonal} in cell free DNA (cfDNA) based cancer detection. It got almost everything right, but said that CH could cause leukemia syndromes. This is rarely the case in reality. Also if you ask ChatGPT to provide relevant references of its provided answer, it will dodge this question by saying: \textit{``I don't have direct access to current databases or research papers ...''} and will instead provide some possible key words to search with (Record S4). Finally, if you ask question about some specific statistics, it will avoid giving any specific numbers. For example, we asked how often copy number variation (CNV) in cfDNA is position specific. ChatGPT simply talks about cases where CNV can be position specific (better than providing false information) (Record S5).

\section{Computational Biology Tool Searching}
When we need to perform computational tasks related to bioinformatics (protein structure prediction, protein similarity searching, genome realignment etc.), the first question that comes to our mind is: \textit{``Is there any tool available for this task already?''} One would most likely ask this question to someone more experienced, because Google searching tool availability returns some disorganized contents. For example, if you ask Google to name some useful tools used for similarity based DNA/protein sequence filtering, Google will give you the link of Uniprot database and some algorithm blogs; this is not very useful. On that other hand, ChatGPT will provide you with short descriptions along with advantages and disadvantages of nine popular relevant tools including BLAST  \cite{altschul1997gapped}, CD-hit \cite{li2006cd} and USEARCH \cite{edgar2010search} (Record S6). Similarly, we asked for a list of popular tools used for mutation variant allele frequency (VAF) calling to ChatGPT. It provided 12 such callers along with description that include VarScan \cite{koboldt2009varscan}, Mutect2 \cite{cibulskis2013sensitive} and Strelka \cite{saunders2012strelka,kim2018strelka2} (Record S7). Finally, we asked for tools that can be used for protein 3D structure prediction from the corresponding 1D sequence. ChatGPT listed out 10 relevant tools including SWISS-MODEL \cite{waterhouse2018swiss}, I-TASSER \cite{yang2015tasser} and ROSETTA \cite{baek2021accurate} (Record S8). It also mentioned the type of approach each tool uses; for example, SWISS-MODEL uses homology modeling while I-TASSER and ROSETTA perform de novo structure prediction. Our investigation convinced us that for searching bioinformatics tools and relevant information, ChatGPT is better than using typical search engines. Hence Google has declared ``code red'' due to the threat of ChatGPT to its searching service.

\section{Bioinformatics Algorithms Made Easy}
Computational biologists use different software tools and built-in libraries to implement Bioinformatics algorithms. One needs to have a basic understanding of the algorithms they are using to use them effectively. If you ask ChatGPT to describe BLAST search to you as if you were a 5-year-old, then ChatGPT will describe the search using pictures and puzzles, which is really easy to understand. Similarly, we asked ChatGPT about explaining the difference between PCA (principal component analysis) \cite{shlens2014tutorial} and Robust PCA \cite{candes2011robust} to a 10-year-old. Although this is not the simplest of topics, ChatGPT explained it using balls of different colors and shapes (Record S9). The explanation was spot on. Finally, we asked the tool to describe us the different flags (used for read filtering) that are seen in the BAM files (a specific bioinformatics file format) It listed out all the flags and explained every one of them perfectly (Record S10).

\section{ChatGPT as Coding Assistant}
Many researchers/developers coming into computational biology belong to diverse backgrounds (including non-programming backgrounds). 90\% of such people require using computational methods at a regular basis \cite{barone2017unmet}. Bioinformatics tools, software and code bases are not nearly as well documented and user friendly as some other sub-fields of computer science. We test how well ChatGPT can support relevant programming tasks. We start by asking ChatGPT to write code for realigning a BAM file to a new reference genome. It provided a step-by-step explained code in pysam. We asked it to do the same thing using SAMtools and BWA \cite{li2009fast}. ChatGPT again produced well documented code along with instructions on how to install these two software (Record S11). We then asked ChatGPT to count the number of DNA sequences inside a fasta file, where it provided a simple one line bash command (Record S12). ChatGPT is great at generating correct bash commands even for complicated tasks. Bash syntax is relatively difficult for less experienced people. Finally, we asked ChatGPT to write a python code that BLAST searches a protein against a local database. It provided a well explained Biopython \cite{cock2009biopython} code (Record S13).

If you ask ChatGPT about library or implementation that does not exist, sometimes, it may give you made-up answers. We told ChatGPT to provide us with sklearn  \cite{scikit-learn} library code for robust PCA. It let us know that no such implementation exists, but provided us with some made-up code from a made-up library called r\_pca (Record S14). Now if you ask it to use sklearn to align fastq pairs to a reference genome, it will correctly let you know that it is not possible. We also asked ChatGPT to give us some code that could calculate edit distance between a pair of sequences in O(n) time. It correctly identified that it is not possible and instead provided us with a code that takes O($n^2$) time (Record S15). We can use ChatGPT for coding assistance by ensuring that we test out the code that it produces.

ChatGPT may lack real world knowledge and critical thinking while giving answers \cite{hanna2023and}. We instructed ChatGPT to write code for generating a random position within the GRCh37 reference genome. It selected a chromosome first and then selected a random position from that chromosome. It did not realize that the chromosomes vary quite a lot in terms of their size and so, the code that it produced would over represent small sized chromosomes (Record S16). We again asked ChatGPT to write code for comparing five groups of values using t-test. It used python scipy \cite{virtanen2020scipy} library to do this in a pair-wise fashion and did not realize that it should use Anova test for such cases instead (Record S17). Nevertheless, ChatGPT does a fine job if the real world knowledge is relatively common. For example, we asked it to calculate the mean targeted sequencing depth of a BAM file obtained from deep targeted sequencing. It correctly calculated number of reads for only the targeted regions using a sample bed file and divided the total number of reads by the total targeted region covered area; this shows its ability to differentiate between WGS and targeted sequencing (Record S18).

\section{Data Visualization}
Data visualization enables adoption of appropriate feature extraction techniques and improves data interpretability. If you have some criteria for a plot but are not sure of what kind of plot should be used, then ChatGPT can help you. We asked ChatGPT the following question: \textit{``I want a plot where x-axis contains continuous values and Y axis contains categorical variables. Each point of the plot belongs to a particular categorical variable of the Y-axis and holds the value of X-axis. For each category, the point also belongs to a particular class and we need to give a different color accordingly. What type of plot should I use?''}. It suggested us to use swarm plot and provided some basic codes to get started. It was correct in interpreting our desired plot (Record S19). Plotting can be complicated because of code syntax. ChatGPT can quickly get you started with complex plots. We gave ChatGPT a list of genes, their significance values and their fold change values. We also mentioned that the genes were coming from two different sources. We asked it to write a python code to generate a volcano plot from these given data. ChatGPT generated relevant code but forgot to perform log for the p-values of y axis and did not label the points with gene names. Mentioning these matters to ChatGPT in the same thread helped it produce the correct code (Record S20). We need to be cautious of these small details while plotting with ChatGPT. Finally, ChatGPT can improve plot visualization. We generated line plots for 19 mutation VAFs (one line plot per mutation) for patient progress monitoring. It was difficult to maintain good color contrast among these 19 line plots plotted together. So, we asked ChatGPT to provide appropriate contrastive color codes from python seaborn library \footnote{https://seaborn.pydata.org/} color palette and obtained the appropriate codes instantly (Record S21).

One problem with ChatGPT is - it will try to generate a plot even if nothing as such exists. We asked it for some R code to visualize an anchor plot (does not exist). It calls a made-up function named AnchorPlot from Seurat R package \cite{satija2015spatial} with nice documentation (Record S22). We also asked ChatGPT for some python code to visualize a BAM-SAM plot (these are sequencing file formats). It assumed that we were talking about \textit{``Box-And-Whisker Scatterplot''} used for comparing data distribution across multiple groups (Record S23). In practice, people never ask ChatGPT to make such random plots and so, it should not be a pressing issue.

\section{Code Conversion}
During customization of algorithms and manipulation of data at a more granular level, conversion of high level library code to a low level language is necessary. For example, one can use pysam for performing more granular selection and feature extraction in BAM files compared to high level samtools library. We asked ChatGPT to convert a SAMtools code that performed preprocessing of BAM files and selection of reads to pysam code. It could do this accurately with proper documentation (Record S24). Going through online pysam documentations to do the same thing would have required a lot of time. We further told ChatGPT to convert a deepTools library \cite{ramirez2014deeptools} based coverage calculation code to R code. Initially it provided us with a made-up function for \textit{``csaw''} library of R. When we prompted again with \textit{``no such library exists''}, it gave the correct code through \textit{``rtracklayer''} library based ``Wig'' file generation (Record S25). This situation further emphasizes the importance of testing out ChatGPT provided code and skimming through the documentation of the provided functions. It is helpful for ChatGPT if you provide it with the library, language name and context of the code that you are trying to convert; but it is not absolutely necessary. We asked ChatGPT to transform an R line plot code to python code without mentioning the programming language or the context of the code. It successfully converted the code (Record S26).

 Code conversion also involves making codes more runtime efficient. We had around a thousand long sequences, where we needed to find the edit distance between each possible pair. ChatGPT converted our python code to C code (python is around 100X slower compared C) successfully when asked (Record S27). Let us look at another example. We had a list containing chromosome positions and another list containing blacklisted regions (each position stored as an integer). We needed to exclude out the blacklisted positions from the original chromosome positions. It would take O($n^2$) time using lists and nested loops. ChatGPT produced a more efficient version of this code by converting the lists to sets (Record S28). Although set look up in python takes O(1) time, this is efficient only when there are not too many positions spanned by the blacklist. If the regions are large, keeping each position as a distinct entry in the set places a large demand on memory. A simpler and more efficient algorithm, assuming the regions in the blacklist are lexicographically sorted according to their start and end points, is to perform a Synchrony join \cite{perna2022iterating}, which is akin to a merge, of the two lists.

\section{Creating Pipelines}
Creating pipelines can assist in (1) streamlining data processing and analysis of large sequencing data, (2) reproducing the multiple steps of a process by collaborators, (3) efficient reuse of the setup process in multiple datasets and experiments, and in (4) integrating multiple software and libraries focused on a single task. We asked ChatGPT to create a pipeline script that (a) uses SAMtools to downsample a BAM file and (b) creates a new BAM file from the output BAM file of step (a) that only has fragments over length 400. We also specified that the output BAM file in step (a) cannot be stored anywhere. ChatGPT used SAMtools and pysam to achieve this but created the intermediate BAM file. When we again specified that the intermediate BAM file could not be created, it then used the concept of piping and here-document to directly send in the streamed downsampled BAM file produced by SAMtools to pysam for further processing without explicitly creating any intermediate BAM file (Record S29). Next we gave ChatGPT a csv file with three columns - (1) cohort names, (2) sample names and (3) tumor fractions - and asked if it could give us a piped bash command that (1) keeps rows where cohort names contain only numeric characters, (2) removes rows with less than 10\% tumor fraction and (3) outputs the filtered csv. We did not specify the data types, but ChatGPT provided the correct pipeline using \textit{``grep''} and \textit{``awk''} (Record S30).  Finally, we instructed ChatGPT to produce a variant calling pipeline using SnakeMake \cite{koster2012snakemake}. It successfully did the following: (a) described what variant calling is, (b) helped in setting up the environment, (c) provided pictorial demonstration and (d) finished its answer with running instructions (Record S31). This can be a great way for computational biologists to familiarize themselves with such new tools in a practical manner.

\section{Code Explanation and Documentation}
When using open source code or code from your collaborators, it is essential to understand that code in order to use it effectively for your own tasks. The use of different libraries and different languages and frameworks in computational biology make it impossible to be an expert in all of them. ChatGPT can fill in this gap. We asked ChatGPT to explain a pipeline code used for genome realignment and realigned BAM file sorting. ChatGPT provided us with (a) a high level overview of the command, (b) description of what each component of the pipeline does and (c) explanation of each tunable parameter used in the command (Record S32). We then told it to explain the \textit{``SeqIO.parse''} function without mentioning the function source. It immediately realized that this function was from Biopython and gave us a sample code (Record S33). One of the best features of ChatGPT is its ability to explain regular expressions which are used a lot during different file processing and filtering. We asked ChatGPT to explain the following regular expression: ``\texttt{re.search(r'(.*\_R)[1|2](\_.*)', fq, re.I|re.M)}''. It first explained every component of this expression correctly and concluded with what type of pattern this expression is intended to look for (Record S34). 

It is best not to ask ChatGPT to explain codes that are long (more than 100 lines), because it will fail to give you an overall sound understanding of what is happening in the code. Most of these large code bases, if created by research folks or PhD students, do not follow industry-level best practices \cite{trisovic2022large}. If you already know what a long code base is supposed to do, then you can ask ChatGPT for a more granular explanation of each code block to understand the code better. We asked ChatGPT to explain a code (around 200 lines) that takes a directory of samples (one sample is one folder with many fastq files) as input and makes sure that each sample has only one fastq pair (paired end sequencing) by combining the fastq files per sample. It provides a high level explanation of each code snippet which is not very clear for a good understanding of the entire code base. When we asked it to give line-by-line comment in the code, it then commented in the entire code and explained the code snippets in a more granular fashion (Record S35). One word of caution is that ChatGPT has a token limit of 4096 while generating answers. When we provided ChatGPT with a 400 line code for providing comments, ChatGPT stopped midway because of token limit. We then prompted ChatGPT in the same thread by saying, \textit{``It is not really the full code''}. Then it commented in the rest of the code (Record S36).

\section{Chatting, Clarifying and Debugging}
Large language models such as ChatGPT are proficient in natural language processing, but they cannot execute code that they generate and so, they often require human feedback on the generated code  \cite{austin2021program}. Piccolo et al. tested ChatGPT with 184 programming exercises from an introductory level bioinformatics course. On the first attempt, ChatGPT solved around 75\% of the exercises. For the remaining exercises, the authors provided natural-language feedback to the model, prompting it to try different approaches. Within 7 or fewer attempts, ChatGPT solved over 97\% of the exercises \cite{piccolo2023many}, which shows the importance of conversations with ChatGPT for obtaining correct code.

A sample conversation with ChatGPT regarding BAM file flags may have the following sequence of questions: \textit{``(1) Explain the flags of a bam file, (2) What is 0X in these flag numbers? (3) How can I check if a read satisfies a certain flag?''}. ChatGPT keeps track of the entire context of the chat thread and can thus answer relevant questions better if asked in the same thread. Another example can be regarding fasta file: \textit{``(1) How do I create a fasta file from a python list of strings? (2) What does ``enumerate'' do in the code you have just provided? (3) How do I merge multiple fasta files into one fasta file?''} (Record S37). If ChatGPT does not understand a particular question, having a conversation in the same thread is a good way for clarification. We asked ChatGPT about the term ``card'' of a matrix found in one the robust PCA papers. ChatGPT was confused. So, we gave the entire formula of the paper where this term was used. ChatGPT was now able to correctly understand that ``card'' is nothing but matrix cardinality (Record S38). If the code generated by ChatGPT throws some error, having further conversations with it about the error is probably the best way in dealing with the error. We told ChatGPT to provide us with some code to filter out variants with less than 50 depth from a bgzipped vcf file. It used bcftools \cite{danecek2021twelve}. The provided code would be correct for most variant callers. But in our case, the system threw \textit{``ambiguos filtering expression error''}. When we asked ChatGPT about this issue in the same thread, it immediately understood that this error was caused by the multiple types of ``DP'' fields created by Mutect2 \cite{cibulskis2013sensitive} variant caller and resolved the issue (Record S39).

Even if you simply specify your error in natural language to ChatGPT without having a conversation about any particular code, it can still resolve the issue quite often. In one case, while trying to generate a plot in the form of a PDF using python matplotlib, we were always getting an empty PDF. Failing to find a solution via Google search, we asked about this issue to ChatGPT. It suggested us to change the position of the \textit{``plt.show()''} command and it worked (Record S40). We were also able to resolve compatibility issues related to pandas and seaborn library during plotting using ChatGPT, where answers provided in stack overflow \footnote{https://stackoverflow.com/} were not effective (Record S41). Sobania et al. \cite{sobania2023analysis} evaluated ChatGPT on the famous bug fixing dataset QuixBugs. The authors showed that its bug fixing performance is comparable to CoCoNut \cite{lutellier2020coconut} and Codex \cite{chen2021evaluating} designed specifically for such coding specific tasks. The authors mentioned that more information about the error through ChatGPT's dialogue response system increased this success rate even further.

\section{Key Takeaways}
While using ChatGPT for computational biology coding assistance, one should take note of the following points:
\begin{itemize}
    \item The more information (language name, library name, code context) you give ChatGPT, the higher is the probability that it will provide you with the correct code. Prompt engineering used for such automatic code development can increase coding efficiency \cite{white2023chatgpt}.
    \item If you get an error by pasting some code written by ChatGPT, keep asking it about those errors in the same conversation and you should eventually get the correct answer.
    \item Having a good understanding of the algorithm or process you are writing code for is important for using ChatGPT effectively because of three main reasons: (1) it helps you identify bugs generated by the chatbot, (2) it enables you to provide more information and context about the code to be generated and (3) it enables you to incorporate the generalized code generated by ChatGPT in your personal code base.
    \item ChatGPT can lack critical thinking while giving the answers and computational biologists have to sometimes modify the code to get the correct logic \cite{borji2023categorical}.
    \item Since the knowledge base of ChatGPT is fixed and does not evolve every month, it is wise not to ask it about very recent knowledge type questions.
    \item If you are not familiar with a library that ChatGPT is using, definitely verify the information using one of the contemporary search engines, because there is a slight chance of it being made-up. Using generative AI models for giving answers is banned in stack overflow currently because of the possibility of misinformation. Even wrong code generated by ChatGPT comes with comments which make the code look well documented, leading to misplaced confidence in the code.
    \item ChatGPT can introduce small false details in technical descriptions which is often difficult to identify. Do not ask ChatGPT pinpointed descriptive questions that ask for yes/no answer or specific statistics or research paper reference; most likely ChatGPT will just avoid being specific. So, it is not a good idea to use ChatGPT for writing technical papers.
    \item ChatGPT has a tendency of writing bioinformatics codes in python, it even uses python subprocess module for creating scripts for pipelining instead of using bash files. Scripting languages such as python enable researchers to focus more on their high level goals by providing rich libraries and packages \cite{ekmekci2016introduction}.
    \item ChatGPT does not care whether you are a beginner or an advanced programmer while generating code; it just provides the code with the language and the framework that it deems the most appropriate \cite{piccolo2023many}.
\end{itemize}

\section{Concluding Remarks}
ChatGPT can go a long way in helping computational biologists with programming. We need to be open to such new technologies while being cautious of their shortcomings. 
The first large language models fine-tuned to generate code were OpenAI’s Codex \cite{chen2021evaluating} and DeepMind’s AlphaCode \cite{li2022competition}. Codex is currently available via GitHub's Copilot and a plugin for the Visual Studio Code editor. While these tools can only generate code from description, ChatGPT is more general purpose and user friendly. Recently, GPT-4 has been released by openAI on March 2023 and is available as a paid version named ChatGPT Plus for general users. This tool can work on image input data as well and can generate response accordingly. Beam et al. compared the performance of students vs GPT-3.5 and GPT-4 in clinical reasoning final examinations (neonatal board exam given to first and second year students at Stanford School of Medicine). The authors showed that GPT-4 outperformed first and second year students and had significant improvement over GPT-3.5 \cite{beam2023performance}. In spite of such developments, the necessity of human intervention for dealing with biological and clinical data will remain in foreseeable future. In the extreme case, computational biologists expert at utilizing such generative AI models will replace computational biologists who do not know how to use them.

\section{Records of Interactions with ChatGPT}
Here are the links to the transcripts of our interactions with ChatGPT: \\
\head{S1}
\href{https://chat.openai.com/share/b25c771d-1889-4d12-9426-26cc8af80dee}{SAMtools Describe} \\
\href{https://chat.openai.com/share/dc11f4a5-a1fd-40d4-a23d-76acdaf797c1}{Pysam vs SAMtools: Comparison} \\
\\
\head{S2}
\href{https://chat.openai.com/share/d9da4fed-3d41-41e3-b948-ae1460abcfb7}{SWISS-PROT Database} \\
\\
\head{S3}
\href{https://chat.openai.com/share/4f537610-bd18-4f7d-84eb-15b0dc2ae2e3}{WGS Cost in 2021} \\
\href{https://chat.openai.com/share/8d767fec-9d18-4ee7-9ab4-07020033d1b0}{Latest PSI-BLAST Version} \\
\\
\head{S4}
\href{https://chat.openai.com/share/6b74746b-7f96-46de-abc7-ec3c2ae5f404}{Clonal Hematopoiesis in cfDNA} \\
\\
\head{S5}
\href{https://chat.openai.com/share/2340ad07-4a07-4f56-9ead-ccc101b97ab0}{cfDNA Copy Number Variation Frequency} \\
\\
\head{S6}
\href{https://chat.openai.com/share/7f3a375c-9c18-478e-9e57-21b2bdbacbac}{Sequence Similarity Analysis Tools}\\
\\
\head{S7}
\href{https://chat.openai.com/share/34917d8d-3969-4ba5-a424-ea1a9c6a528f}{Popular Mutation VAF Calling Tools}\\
\\
\head{S8}
\href{https://chat.openai.com/share/4867ed08-16d9-4d1a-a084-c162db3b7b9f}{Protein 3D Structure Prediction}\\
\\
\head{S9}
\href{https://chat.openai.com/share/db62b56a-2a19-48ed-ae26-f565ffe7747a}{Explaining BLAST Search} \\
\href{https://chat.openai.com/share/6a88f0d6-1202-47b0-b18d-253c4ecc7365}{PCA vs Robust PCA} \\
\\
\head{S10}
\href{https://chat.openai.com/share/ff53ba8a-de43-49dd-b17b-d5a4c4b9abcf}{BAM File Flags Explanation} \\
\\
\head{S11}
\href{https://chat.openai.com/share/95b093cc-bfa7-4161-9020-0d51cc529a17}{Realign BAM to Genome}\\
\\
\head{S12}
\href{https://chat.openai.com/share/8d301b12-3ac7-40d7-8ea2-8c1e02b8218c}{Counting FASTA Sequences: Bash}\\
\\
\head{S13}
\href{https://chat.openai.com/share/793a2021-fe99-41bd-bdaa-65539ab5ef42}{Python BLAST Protein Search}\\
\\
\head{S14}
\href{https://chat.openai.com/share/ecfc4358-77ba-4375-9428-b04d676b8a1c}{RPCA using rpy2 in Python}\\
\\
\head{S15}
\href{https://chat.openai.com/share/a17a8d69-dd99-441d-b0a2-dd8c428d60c6}{Edit Distance Calculation Code}\\
\\
\head{S16}
\href{https://chat.openai.com/share/f62b578d-ef0b-41a8-bdcc-5a52a7218057}{Random GRCh37 Genome Position}\\
\\
\head{S17}
\href{https://chat.openai.com/share/f07f6746-3af3-4eee-8d5f-1afea09de055}{T-Test for Group Comparison}\\
\\
\head{S18}
\href{https://chat.openai.com/share/37899080-9033-4190-a2d1-f49283c3ea8e}{Mean Targeted Seq Depth Calc}\\
\\
\head{S19}
\href{https://chat.openai.com/share/4c4a9005-5c5e-406e-9bf7-ed6a2f9acf15}{Swarm Plot}\\
\\
\head{S20}
\href{https://chat.openai.com/share/18f44ab6-9a92-4e20-bd7a-653d7837c405}{Volcano Plot Python Code}\\
\\
\head{S21}
\href{https://chat.openai.com/share/1e8f4f2f-0afd-408a-825c-cbbf24b167d3}{Contrastive Colors for Lineplot}\\
\\
\head{S22}
\href{https://chat.openai.com/share/a1bfb479-b823-42bb-b842-c11c00d14375}{Anchor Plot in R}\\
\\
\head{S23}
\href{https://chat.openai.com/share/2c91afdd-ebc2-4d30-9f92-c6401c013b48}{BAM-SAM Plot}\\
\\
\head{S24}
\href{https://chat.openai.com/share/fafaa061-f200-447a-be8a-8dcb687ba03f}{BAM File Preprocessing in PySam}\\
\\
\head{S25}
\href{https://chat.openai.com/share/ba3a520b-256f-4486-ae45-5794cea43d15}{R Code for Coverage Calculation}\\
\\
\head{S26}
\href{https://chat.openai.com/share/5690d7e3-72f6-41ee-928b-b2762f71b935}{R-to-Python ggplot}\\
\\
\head{S27}
\href{https://chat.openai.com/share/b086f9a6-fce1-4a31-b02e-b8fcff3cdacf}{Python-to-C Edit Distance}\\
\\
\head{S28}
\href{https://chat.openai.com/share/1e3f8829-928b-4c40-8556-c0e291d6ed45}{Exclude Blacklisted Positions Efficiently}\\
\\
\head{S29}
\href{https://chat.openai.com/share/46fa168c-6bb6-41dd-bfc4-fe55a50696d0}{Creating Pipeline for Downsampling and Fragment Selection}\\
\\
\head{S30}
\href{https://chat.openai.com/share/8f63913a-4e85-4c12-bbbd-0861ba9a5805}{Filtering CSV with Bash}\\
\\
\head{S31}
\href{https://chat.openai.com/share/2dbf5c76-a509-4754-9b02-c74d34fae1da}{SnakeMake Variant Calling Pipeline}\\
\\
\head{S32}
\href{https://chat.openai.com/share/ab79ab02-426e-4173-8e1b-f32033961a02}{BWA Alignment and Sorting Explained}\\
\\
\head{S33}
\href{https://chat.openai.com/share/a157e2bb-58c9-4c6d-acba-7f9dae1ebd22}{Parse FASTA into list}\\
\\
\head{S34}
\href{https://chat.openai.com/share/5f6cb1c1-148d-4222-9d22-46db32615afc}{Regular Expression Explanation}\\
\\
\head{S35}
\href{https://chat.openai.com/share/5f6c367a-bc5a-4be8-9531-f965f8dda521}{Long Code: Merge FASTQ Files Script}\\
\\
\head{S36}
\href{https://chat.openai.com/share/bab32294-2795-4b25-be01-a47bc9bae058}{Documenting bcbio Python code}\\
\\
\head{S37}
\href{https://chat.openai.com/share/225814c0-047c-4527-bbb4-5461dd23e400}{Bam Flag Thread}\\
\href{https://chat.openai.com/share/d7dd4616-2e05-46ca-b278-62de06ec733a}{FASTA File Thread}\\
\\
\head{S38}
\href{https://chat.openai.com/share/971f73c1-5fd7-4c46-a946-5d2518ec8e51}{Matrix Card Thread}\\
\\
\head{S39}
\href{https://chat.openai.com/share/d6ae7692-b92e-4692-ad92-6931af5bd127}{Filter VCF by Depth}\\
\\
\head{S40}
\href{https://chat.openai.com/share/70cb6356-f9cc-4eee-a8bb-3d44c9552b03}{Troubleshoot PDF Plotting Issues}\\
\\
\head{S41}
\href{https://chat.openai.com/share/deba0d52-9889-45bd-ae61-6ad7d3e1a513}{Resolve Seaborn Option Error}\\
\\

\bibliographystyle{splncs04}

\bibliography{main}

\end{document}